\documentclass[letterpaper, 10 pt, conference]{ieeeconf}  

\IEEEoverridecommandlockouts                              
\usepackage[utf8]{inputenc}
\usepackage[T1]{fontenc}
\usepackage[noadjust]{cite}

\usepackage{graphicx} 

\title{\LARGE \bf
A different take on the best-first game tree pruning algorithms
}

\author{Ishan Srivastava$^{1}$
\thanks{*This work was not supported by any organization or any other person. This is independent research performed during the time the author was pursuing their undergraduate studies, not supervised by any faculty member.}
\thanks{$^{1}$Ishan Srivastava is currently a B.Tech. CSE undergrad student at Indian Institute of Technology, Dharwad, 580011, Karnataka, India
        {\tt\small ishan.pr.srt@gmail.com}}%
}

\begin{document}

\maketitle
\thispagestyle{plain}
\pagestyle{plain}

\begin{abstract}

The alpha-beta pruning algorithms have been popular in game tree searching ever since they were discovered. Numerous enhancements are proposed in literature and it is often overwhelming as to which would be the best for implementation. A certain enhancement can take far too long to fine tune its hyper parameters or to decide whether it is going to not make much of a difference due to the memory limitations. On the other hand are the best first pruning techniques, mostly the counterparts of the infamous SSS* algorithm, the algorithm which proved out to be disruptive at the time of its discovery but gradually became outcast as being too memory intensive and having a higher time complexity. Later research doesn't see the best first approaches to be completely different from the depth first based enhancements but both seem to be transitionary in the sense that a best first approach could be looked as a depth first approach with a certain set of enhancements and with the growing power of the computers, SSS* didn't seem to be as taxing on the memory either. Even so, there seems to be quite difficulty in understanding the nature of the SSS* algorithm, why it does what it does and it being termed as being too complex to fathom, visualize and understand on an intellectual level. This article tries to bridge this gap and provide some experimental results comparing the two with the most promising advances.

\end{abstract}

\section{INTRODUCTION}

The existence, invention and re-invention of alpha-beta pruning methods for game trees can be traced back to as far as 1956 \cite{wiki-alpha-beta}. The algorithm seemed to have been invented simultaneously independently across continents and immediately became a favourite for the traversal of game trees in famous Turn Based Strategy Games like chess, checkers, Othello-reversi etc. The 19th century witnessed a number of variations and wide variety of enhancements, research and benchmarks formulated for these pruning techniques. Another algorithm SSS* proposed by Stockman in 1979 \cite{stockman_1979} proved out to be disruptive. Having a totally different approach to the problem of game trees, being a state space search algorithm instead of traversing the tree in a depth-first manner, having experimental results and theoretical proof that SSS* never explores a node that the alpha-beta ignores, yet having high memory requirements led to its quick popularity and eventual downfall, finally being proclaimed as dead at the \textit{8th Advances in Computer Chess conference 1996} \cite{SSS-wiki}.

Still, yet to date the SSS* algorithm continues to excite the population, being a crucial part of textbooks and curriculum and even though termed by many as too complicated to understand, not-intuitive or complex \cite{stockman_no_yes} and slower with regards to the newer advances in alpha-beta pruning methods \cite{slow_1}\cite{slow_2}, it is hard to let go unnoticed, the brilliance of the backbones on which it was based upon. While everyone was improving the depth-first method, SSS* looked at the problem in a best first manner, more so it searched in the solution space of “strategies” which this article explores in great detail, this aspect being neglected in previous literature. The SSS* was termed as difficult to understand as it was shockingly different from the research being pursued in the field at that time but it would be wrong to say that it was non-intuitive.

The human brain often correlates things from different domains when coming up with a solution to a problem and SSS* has definitely earned its place in the syllabus of any course/textbook on AI

We will explore why there still exists confusion between developers whether to go for the SSS*, we provide experimental results on to whether SSS* can still compete with the other advances in alpha-beta pruning, regarding the number of nodes explored which can be beneficial in specific scenarios where the node evaluations is done in a constant negligible amount of time or when it requires to be done again and again in games where the heuristic value of a board configuration can change with respect to the time elapsed in the game. We finally reiterate the fact that both algorithms are not different but one can be proved to be the other one with change in certain enhancements.

\section{THE SSS* ALGORITHM}
The initial development of the alpha-beta algorithm was in a sense that it was a blind algorithm and searched the tree from left to right. It was evident that the algorithm can be improved provided that the better nodes were visited first or were found on the left part of the tree. Efforts were made to improve this move reordering using iterative deepening and transposition tables combined with various other enhancements like the history heuristic, minimal window, aspiration search, variable search depth, killer heuristic etc. \cite{ref1}\cite{ref2}\cite{ref3}. SSS* employed the brute force way to order these moves, we later see that there isn’t that great a difference between depth first alpha beta and best first SSS*. There is a smooth transitioning between the two achieved by the various enhancements to the alpha beta and SSS* can be ultimately formulated as a special case of repeated null window alpha-beta calls with transposition tables \cite{ab_is_sss}.

SSS* was one of the very first attempts at an algorithm that wasn’t uninformed and which had a sense of direction. As the parameters for variable search depth aren't quantifiable we don't talk about that much in this article. Variable search depth is achieved during iterative deepening where we search to a greater depth for the moves that gave out better values during the last iteration.

\subsection{Strategy}
We begin by defining the notion of a strategy\cite{youtube}. A strategy can be thought of as a set of moves that the MAX player has thought of playing for every possible MIN move after searching K-PLY deep into the game tree from the current configuration/state. And this is generally how new players to a TBS game start out playing. They are unaware of the recognisable board configurations that more expert players try to recall which might lead to a quicker/easier win or a loss, rather they start out by thinking about the possible scenarios that might occur for every MIN move and have a response in mind to each and every one of it, as an agent not having any domain specific knowledge. Whatever may be the MIN’s move, MAX has an answer prepared for that. Such a subtree is called a strategy. 

The SSS* searches in the state space of strategies.

\subsection{Value for a Strategy}
The Value of a strategy is defined as the minimum of the eval/heuristic values of all the leaf nodes. Because once a strategy is fixed by MAX, the player responsible for driving the outcome is MIN and it will always try to minimise the value of the game tree (assuming a perfect player at the other end).

Consider the subtree in figure 1 \cite{youtube}, with the given eval values. MAX nodes are represented by squares while the MIN nodes are represented by circles. The tree is fairly symmetrical for understanding purposes and has a constant branching factor of 2. We search 4-PLY deep and apply the \textbf{eval()} function at the leaves. The nodes in grey are the ones not explored by vanilla alpha-beta pruning. The eval values are such that the better nodes are distributed more towards the right side of the tree.
\begin{figure}[htb]
  \centering
  \includegraphics[width=0.45\textwidth]{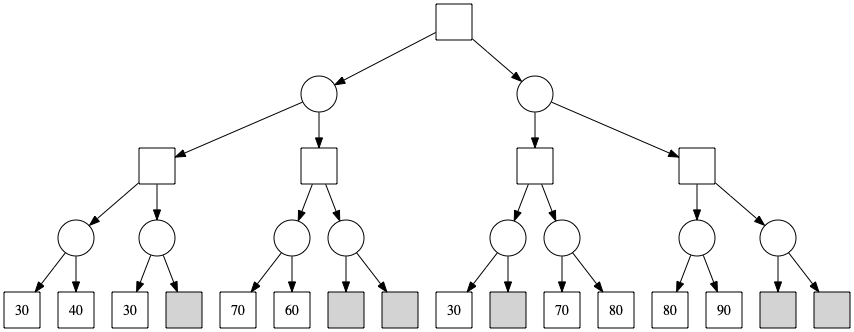}
  \caption{Example Game Tree}
\end{figure}

\subsection{Clusters}
A leaf node may be a part of more than one strategy. In this particular example every leaf belongs to 2 strategies which can be visualised by forming the strategy subtrees considering all choices for MIN and one choice for MAX. The SSS* is exhaustive in the sense that it never misses out on the best strategy. This sounds more like brute force but that isn't what is actually performed by SSS*.

The relation between the value of a leaf node L and the value of the strategy S can be defined as that \textit{any value L will be an upper bound on the value of the strategy S}. This way we can say that a leaf node represents a \textit{partial strategy} i.e. which is not fully solved. The leaf nodes represent more than one strategy which we will refer to by its \textit{cluster of strategies}. Whatever might be the value of the leaf node L, it will be an upper bound on all the strategies belonging to that cluster.

We maintain a priority queue of clusters and clusters are not represented explicitly, but essentially by their representative leaf node. Each element in the priority queue is represented by a 3 tuple comprising of the name of the node, the notion of whether it's LIVE or SOLVED and an upper bound on the eval value: \textbf{\textit{<name, live/solved, bound>}}.
The highest value will always be at the head of the queue, so the priority queue is a max queue with repect to the eval value propogated upwards as the solution of the subtree beneath that node.

To cover all the strategies i.e. obtain a subset of leaves that span all the strategies, constitutes the first part of the SSS* algorithm regarded as its \textit{forward phase}. We can see that this can be obtained intuitively by covering \textit{all branches for the choice of MAX and choosing any one of the choices for MIN}. Following this, it will lead to picking the leaf nodes that span all the strategies. A strategy was obtained by choosing one choice for MAX while considering all choices for MIN. The clusters that span all the strategies can be obtained by doing the \textit{``opposite"} i.e. by choosing all choices for MAX but one choice for MIN.

Now that we have made sure that we aren’t missing out on any strategies, the higher level algorithm for SSS* can be defined as:
\begin{enumerate}
    \item Find the best looking partial strategy
    \item Refine until best solution is found out
\end{enumerate}
since its a best first algorithm.

SSS* maintains a priority queue of the representative nodes. Each element in the priority queue is either LIVE or SOLVED. So when we initially start out, we start with the root node and we mark it as LIVE. The problem seems to be very comparable to the famous AO* algorithm in the sense that we consider only one choice for MIN which can be compared to the OR node and the AND node can be compared to MAX. In the AO* algorithm we had nodes that were solved or not solved. In this case we are explicitly stating them as LIVE. The SSS* algorithm keeps on going until the root node is solved which is another similarity with the AO* algorithm. Both are Best First algorithms.

We start with the root node, mark it as LIVE and put a bound of \textbf{+large} and insert this tuple into the priority queue.

Similar to AO* algorithm, SSS* also has two separate forward and backup phases. In the forward phase we are essentially looking for the clusters to span all the strategies. In Fig 2 these 4 nodes highlighted in pink represent the 4 clusters. Each cluster represents 2 strategies. The figure shows how we have considered all the choices for MAX and arbitrarily left most in this case, single choice for MIN. These 4 leaf nodes represent 2 strategies each. After the terminal nodes have been reached and solved the SSS* proceeds on with the backup phase. These 4 terminal nodes will be sorted out in the priority queue according to their eval value.
\begin{figure}[htb]
  \centering
  \includegraphics[width=0.45\textwidth]{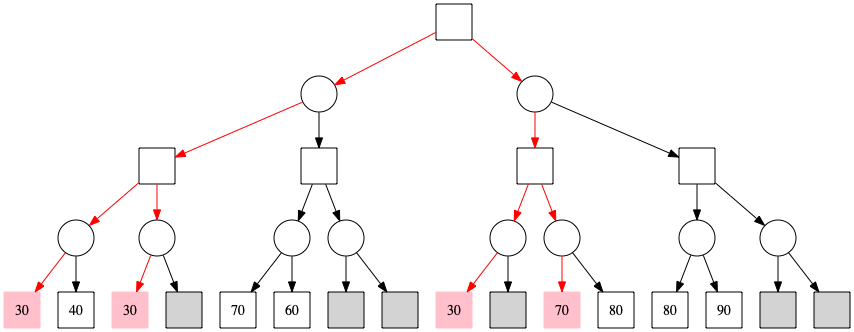}
  \caption{Example game tree indicating the nodes that span all the strategies}
\end{figure}

Without providing the pseudo code for the SSS* algorithm which is freely available we talk about the various cases/conditions and the actions performed during its forward and backup phase which help to understand the algorithm better and typically why SSS* does what it does \cite{youtube}.

\noindent \textbf{forward phase:}
\begin{itemize}
    \item \textit{MAX, live, non-terminal} --> add all children to the queue
    \item \textit{MIN, live, non-terminal} --> add one child
\end{itemize}
\textbf{backup phase:}
\begin{itemize}
    \item \textit{terminal} --> label it solved and put the value of the node as eval values
    \item \textit{MAX, solved} --> replace with sibling of lower value ELSE replace with parent and mark it as solved
    \item \textit{MIN, solved} --> replace with parent and mark it as solved
\end{itemize}

In the example in figure 2, the MAX node with the value 70 would be at the head of the queue and its sibling would be checked. In this case it doesn't have any sibling with a lower eval value in which case we replace it with its parent and label it SOLVED. This is because MIN will be driving the outcomes for the strategy and will always look for minimizing the value of the game tree.

Once we move on to its parent which is a MIN node, there is no reason to check for its sibling or any sibling of any MIN node, since earlier MAX node that we picked at the (current depth + 1) was from the head of the priority queue. Which means the other strategy from the sibling of the current MIN node has an upperbound of a value which is essentially lower than 70 and MAX will never choose this node. So we can label the MIN node as solved and proceed on to its parent. This is somewhat similar to an alpha cut off. At this point the tree looks somewhat demonstrated in figure 3. The nodes visited by the SSS* algorithm are in pink and for now the upper bound on the MAX node at depth 2 is 70. The algorithm now solves for its sibling with an upperbound of 70. The labels in the figure outside the nodes represent the relevant upperbounds.
\begin{figure}[htb]
  \centering
  \includegraphics[width=0.45\textwidth]{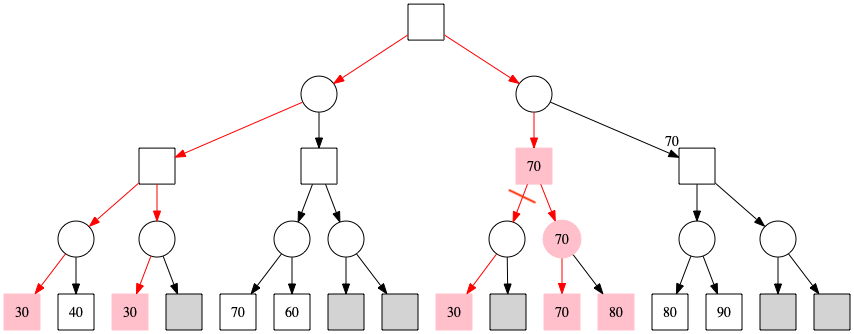}
  \caption{Example Game tree showing the first cutoff for sibling of MIN node}
\end{figure}

Note that this again starts the forward phase of the SSS* for a while until the terminal nodes are reached again. Since the upperbound remains 70 the further nodes in the subtree to be explored under this sibling are solved for either the minimum/maximum of the value of the node or the bound, whichever is lower. The next 2 prominent steps for the execution of the SSS* algorithm are shown in figure 4 and 5.

\begin{figure}[htb]
  \centering
  \includegraphics[width=0.45\textwidth]{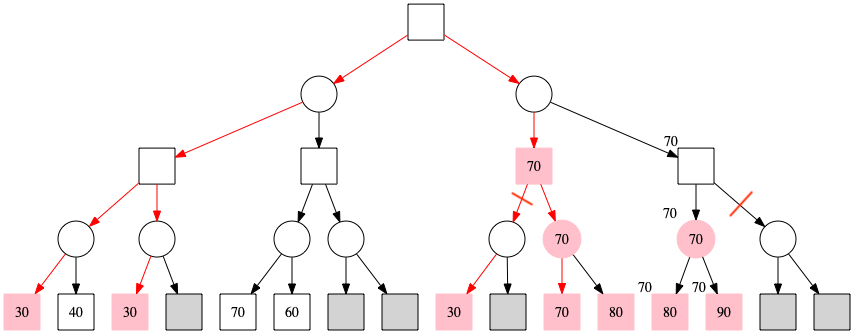}
  \caption{Example Game tree showing a prominent step upon execution of the SSS* algorithm}
\end{figure}

\begin{figure}[htb]
  \centering
  \includegraphics[width=0.45\textwidth]{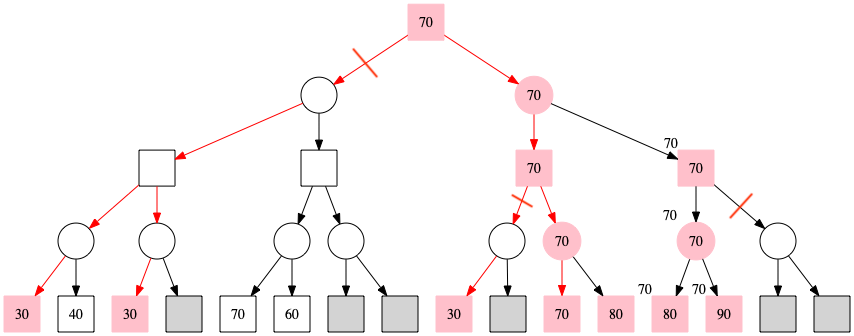}
  \caption{Example game showing the solved solution tree and the nodes explored by the SSS* algorithm}
\end{figure}

It must be noted that the SSS* algorithm focused altogether on the right hand side of the tree, where a better chance of getting the solution was plausible and this example supports the claim by Stockman \cite{stockman_1979} that SSS* never explores a node that alpha beta can ignore and ends up visiting fewer nodes than our vanilla alpha beta. 

\section{RELATED WORK}
Various combinations of advances in alpha beta pruning have already been tried in the past \cite{ref1}\cite{ref2}\cite{ref3}. It has been found that a combination of transposition tables and history heuristic with iterative deepening accounts for almost 99\% of the reduction in game tree size\cite{ref1}. With the help of transposition tables, it is possible to generate trees that are smaller than the minimal tree. Although the definition of minimal tree is fuzzy and for uneven branching factors it is taken as the best case of alpha-beta \cite{minimal_tree}. Some results by Schaeffer show that as the computing power increased and computer memory became cheaper, near 1995 the SSS* algorithm was actually not that inefficient and can actually be implemented for programs like \textit{Phoenix} on the state of the art machines of that time \cite{stockman_no_yes}. However it is inconclusive whether it was taken into consideration shortly after when it was proclaimed as dead.

\section{THE EXPERIMENT}
\subsection{Experimental Setup}
We implement the classic game of Othello-reversi and compare the best combination of enhancements applied to alpha-beta (experimentally proven)\cite{ref1}, with that of SSS* with iterative deepening, dynamic move ordering and transposition tables applied to both. We comapare them in terms of the number of total nodes explored at the terminal level. Although there have been conflicting view points whether the number of terminal nodes explored represent the actual nature of the complexity and the run time of the algorithm, the general view remains that the two are fairly related to each other \cite{ref1} \cite{new_advances}. It is no doubt that it is independent of the system the algorithm is run on and are somewhat associated to the size of the game tree that needs to be explored and the actual time taken for the algorithm to calculate the value of the current state. We average over 50 popular positions picked up by personal preference and which seemed rather common as the historical benchmark positions are regarded as obsolete with no new benchmarks produced for reversi in the recent years \cite{stockman_no_yes}.

\subsection{Results}
Figure 6 shows the number of leaf nodes explore by the SSS* algorithm as compared to the alpha-beta (both with their respective enhancements as described above). As some literature has already disagreed with the earlier view point that $O(w^{d/2})$ is too much memory requirements for practical purposes \cite{stockman_no_yes}, we see that SSS* performs quite close to the alpha-beta and there really isn't much of a difference between the number of leaf nodes explore by those two. The alternating nature for depths in case of the number of leaf nodes explored has also been stated many times in the past, which can be summed up pretty briefly by the fact that the incremental cost of growing the tree an additional PLY to an odd depth is greater than for an even depth \cite{ref1}.
\begin{figure}[htb]
  \centering
  \includegraphics[width=0.45\textwidth]{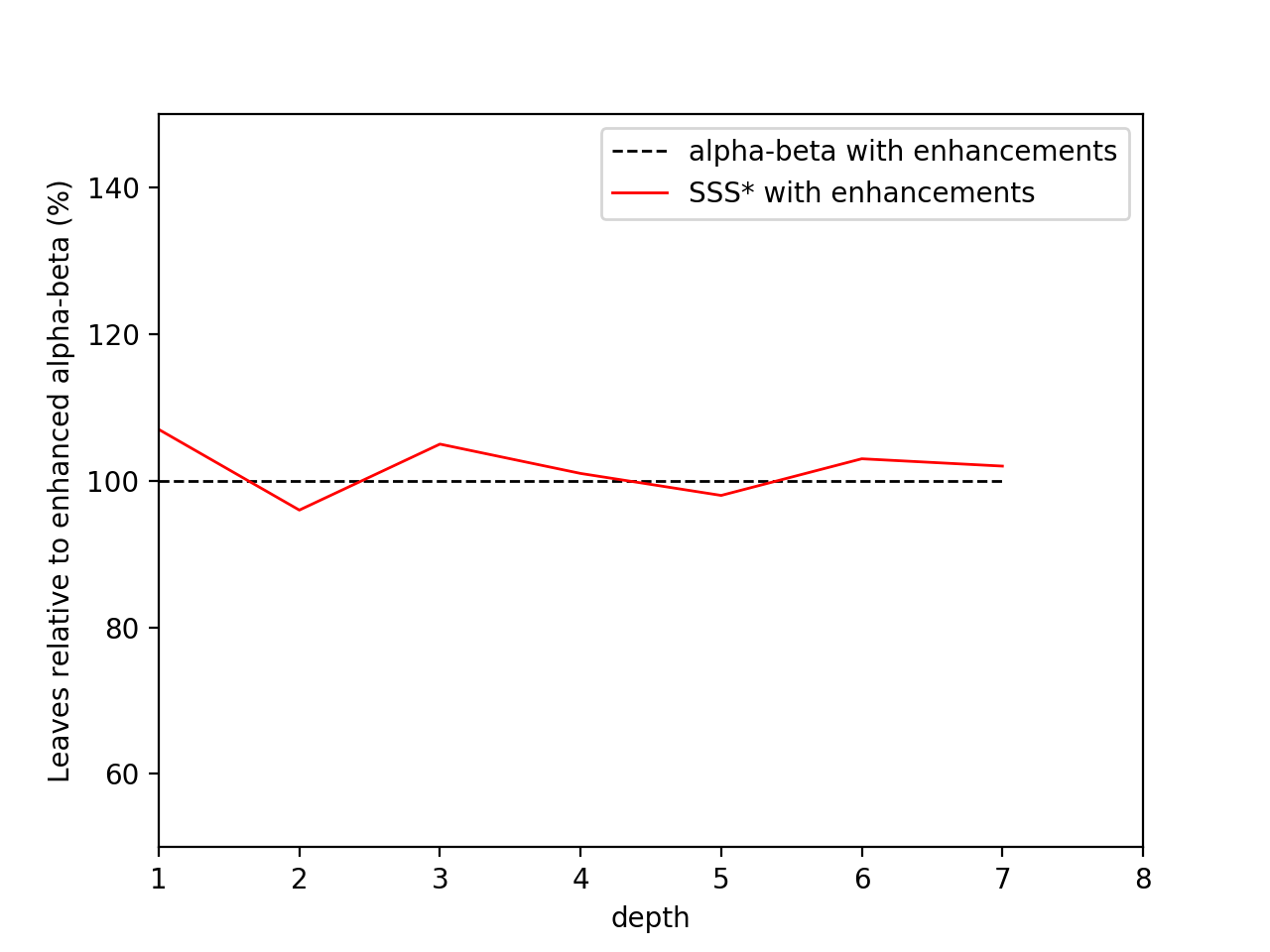}
  \caption{Leaf node count of (both enhanced) alpha-beta and SSS*}
\end{figure}

\section{CONCLUSIONS}

This article was an attempt to bridge the gap between a few game tree exploring algorithms, and a venture to explain the logistics behind the SSS* algorithm. The viewpoint of the SSS* algorithm at the time of its development was the most unconventional and this article tries to explain the missing plots that make it worth having it in any programmer's arsenal. Apart from this, attempts have been made to resolve confusion behind SSS* and not let go unnoticed, the wit that was involved in coming up with this kind of idea at that point in time. As we know now that the two algorithms are very similar in nature and its possible to come up with the other starting from one and changing the hyperparameters of the enhancements used, it has been shown in literature that native SSS* is equivalent to alpha-beta using transposition tables with repeated null window searches \cite{ab_is_sss}. We finally establish the fact that the two can be used interchangeably depending on the use case and it is futile to waste time choosing between the two as both spit out comparable results when used taking care of the various enhancements that have been made out since their discovery.

\addtolength{\textheight}{-12cm}   








\bibliographystyle{ieeetran.bst}
\bibliography{references}

\begin{thebibliography}{10}
\providecommand{\url}[1]{#1}
\csname url@samestyle\endcsname
\providecommand{\newblock}{\relax}
\providecommand{\bibinfo}[2]{#2}
\providecommand{\BIBentrySTDinterwordspacing}{\spaceskip=0pt\relax}
\providecommand{\BIBentryALTinterwordstretchfactor}{4}
\providecommand{\BIBentryALTinterwordspacing}{\spaceskip=\fontdimen2\font plus
\BIBentryALTinterwordstretchfactor\fontdimen3\font minus
  \fontdimen4\font\relax}
\providecommand{\BIBforeignlanguage}[2]{{%
\expandafter\ifx\csname l@#1\endcsname\relax
\typeout{** WARNING: IEEEtran.bst: No hyphenation pattern has been}%
\typeout{** loaded for the language `#1'. Using the pattern for}%
\typeout{** the default language instead.}%
\else
\language=\csname l@#1\endcsname
\fi
#2}}
\providecommand{\BIBdecl}{\relax}
\BIBdecl

\bibitem{wiki-alpha-beta}
``Alpha-beta - chessprogramming wiki - history, quotes and enhancements,''
  \url{https://www.chessprogramming.org/Alpha-Beta}.

\bibitem{stockman_1979}
G.~Stockman, ``\BIBforeignlanguage{en}{A minimax algorithm better than
  alpha-beta?}'' \emph{\BIBforeignlanguage{en}{Artificial Intelligence}},
  vol.~12, no.~2, pp. 179--196, Aug. 1979.

\bibitem{SSS-wiki}
``Sss* and dual* - chessprogramming wiki,''
  \url{https://www.chessprogramming.org/\textunderscore{SSS*\_and\_Dual*}}.

\bibitem{stockman_no_yes}
A.~Plaat, J.~Schaeffer, W.~Pijls, and A.~de~Bruin, ``A {Minimax} {Algorithm}
  {Better} {Than} {Alpha}-beta?: {No} and {Yes},'' Feb. 2017.

\bibitem{slow_1}
A.~Plaat, J.~Schaeffer, W.~Pijls, and A.~Bruin, ``Best-first and depth-first
  minimax search in practice,'' 10 1996.

\bibitem{slow_2}
H.~{Kaindl}, R.~{Shams}, and H.~{Horacek}, ``Minimax search algorithms with and
  without aspiration windows,'' \emph{IEEE Transactions on Pattern Analysis and
  Machine Intelligence}, vol.~13, no.~12, pp. 1225--1235, Dec 1991.

\bibitem{ref1}
J.~Schaeffer, ``The history heuristic and alpha-beta search enhancements in
  practice,'' \emph{IEEE Transactions on Pattern Analysis and Machine
  Intelligence}, vol.~11, no.~11, pp. 1203--1212, Nov. 1989.

\bibitem{ref2}
M.~S. Campbell and T.~Marsland, ``\BIBforeignlanguage{en}{A comparison of
  minimax tree search algorithms},'' \emph{\BIBforeignlanguage{en}{Artificial
  Intelligence}}, vol.~20, no.~4, pp. 347--367, Jul. 1983.

\bibitem{ref3}
T.~Marsland, ``\BIBforeignlanguage{en}{A {Review} of {Game}-{Tree}
  {Pruning}1},'' \emph{\BIBforeignlanguage{en}{ICGA Journal}}, vol.~9, no.~1,
  pp. 3--19, Mar. 1986.

\bibitem{ab_is_sss}
A.~Plaat, J.~Schaeffer, W.~Pijls, and A.~de~Bruin, ``Sss* = alpha-beta +
  {TT},'' \emph{CoRR}, vol. abs/1404.1517, 2014.

\bibitem{youtube}
``Game playing-sss * nptel prof. deepak khemani - youtube,''
  \url{https://www.youtube.com/watch?v=2I9eR3IO6zM\&t=2008s}.

\bibitem{minimal_tree}
D.~E. Knuth and R.~W. Moore, ``An analysis of alpha-beta pruning,''
  \emph{Artificial Intelligence}, vol.~6, no.~4, pp. 293 -- 326, 1975.

\bibitem{new_advances}
J.~Schaeffer and A.~Plaat, ``New advances in {Alpha}-{Beta} searching,'' in
  \emph{In {Proceedings} of the 1996 {ACM} 24th annual conference on {Computer}
  science}.\hskip 1em plus 0.5em minus 0.4em\relax ACM Press, 1996, pp.
  124--130.

\end{thebibliography}

\end{document}